# Gene Expression Time Course Clustering with Countably Infinite Hidden Markov Models

**Matthew J. Beal**  **Praveen Krishnamurthy**
Department of Computer Science and Engineering
State University of New York (SUNY) at Buffalo
Buffalo, NY 14260-2000
{*mbeal,pk35*}*@cse.buffalo.edu*

## Abstract

Most existing approaches to clustering gene expression time course data treat the different time points as independent dimensions and are invariant to permutations, such as reversal, of the experimental time course. Approaches utilizing HMMs have been shown to be helpful in this regard, but are hampered by having to choose model architectures with appropriate complexities. Here we propose for a clustering application an HMM with a countably infinite state space; inference in this model is possible by recasting it in the hierarchical Dirichlet process (HDP) framework (Teh et al. 2006), and hence we call it the HDP-HMM. We show that the infinite model outperforms model selection methods over finite models, and traditional time-independent methods, as measured by a variety of external and internal indices for clustering on two large publicly available data sets. Moreover, we show that the infinite models utilize more hidden states and employ richer architectures (e.g. state-to-state transitions) without the damaging effects of overfitting.

## 1 Introduction

There are a large number of popular techniques for clustering gene expression data, with the goal being to elucidate the many different functional roles of genes that are players in important biological processes. It is said that genes that cluster with similar expression—that is, are *co-expressed*—serve similar functional roles in a process (see, for example, Eisen et al. 1998). Bioinformaticians have more recently had access to sets of time-series measurements of genes' expression over the duration of an experiment, and have desired therefore to learn not just co-expression, but *causal* relationships that may help elucidate *co-regulation* as well.

Two problematic issues hamper practical methods for clustering gene expression time course data: first, if deriving a model-based clustering metric, it is often unclear what the appropriate model complexity should be; second, the current clustering algorithms available cannot handle, and therefore disregard, the temporal information. This usually occurs when constructing a metric for the distance between any two such genes. The common practice for an experiment having $T$ measurements of a gene's expression over time is to consider the expression as positioned in a $T$-dimensional space, and to perform (at worse spherical metric) clustering in that space. The result is that the clustering algorithm is invariant to arbitrary permutations of the time points, which is highly undesirable since we would like to take into account the correlations between all the genes' expression at nearby or adjacent time points.

In i.i.d. data sets, the model complexity issue has been recently and successfully tackled by using Dirichlet process mixture models, and in particular countably infinite Gaussian mixture models (examples can be found in Rasmussen 2000, Wild et al. 2002, Medvedovic and Sivaganesan 2002, Dubey et al. 2004, Medvedovic et al. 2004). However, these models are not applicable to time series data unless ones uses the unsavory concatenation of time points mentioned above (for example, in the work of Medvedovic and Sivaganesan (2002), an i.i.d. infinite mixture of Gaussians is applied to Spellman's time series data). To address this second issue of time-dependence, parametric models such as state-space models and differential equation models have been employed, but the model-complexity issue still needs to be tackled either with heuristic model selection criteria (Iyer et al. 1999), or approximate Bayesian methods such as BIC (Ramoni et al. 2002) or variational Bayes (Beal et al. 2005). For continuous time series, modeling spline parame-ters has been popular (Bar-Joseph et al. 2003, Heard et al. 2006). For discrete time series, which we consider in this paper, random walk models (Wakefield et al. 2003) have been proposed. The most advanced analysis thus far is the mixture of HMMs approach of Schliep et al. (2005), wherein BIC (Schwarz 1978) and an entropic criterion are used for model selection. The approach outlined in this paper also uses an HMM, but we side-step the question of model selection by using a flexible nonparametric Bayesian mixture modeling framework, which allows the model to have a countably infinite number of hidden states (i.e. a countably infinite by countably infinite transition matrix).

We present a simple but powerful temporal extension to the recently introduced Hierarchical Dirichlet Process mixture model (HDP) of Teh, Jordan, Beal and Blei (2004,2005), in the form of a model that is coined the *hierarchical Dirichlet process hidden Markov model* (HDP-HMM). As we describe below, this extension allows us to address both of the issues noted above. Moreover, we can still provide a measure of similarity between any two genes' time courses, by examining the probabilistic degree of overlap in the hidden state trajectories; this is possible despite these state spaces being countably infinite.

The paper is arranged as follows. In Section 2 we briefly review the HDP framework, then show how the countably infinite HMM is a particular (though non-trivial) instantiation within this framework, then describe a straightforward similarity measure for pairs of sequences. In Section 3 we present results of time-course clustering experiments on two publicly available gene data sets, for which ground truth labels are provided, measuring performance with respect to a variety of so-called external and internal indices. We conclude in Section 4 by suggesting directions for future work and expansion of the HDP-HMM.

## 2 The infinite HDP-HMM

In this section we present a hidden Markov model with a countably infinite state space, which we call the *hierarchical Dirichlet process hidden Markov model* (HDP-HMM) by way of its relationship to the HDP framework of Teh et al. (2006). We begin with an overview of the HDP according to this work, and then recast the infinite HMM into this framework. Previous research by the first author on infinite HMMs (Beal et al. 2002) provided an approximate sampling scheme for inference and learning but was unable to prove its correctness. By recasting the infinite HMM as a constrained HDP, we will show that we can have a functioning sampling scheme at our disposal, as explained below.

### 2.1 Hierarchical Dirichlet Processes

The HDP considers problems involving groups of data, where each observation within a group is a draw from a mixture model, and where it is desirable to share mixture components between groups. Consider first a single group of data, $(x_i)_{i=1}^n$; if the number of mixture components is *unknown a priori* and is to be inferred from the data, then it is natural to consider a Dirichlet process mixture model, as depicted in Figure 1(a) (for an exposition on DP mixtures see Neal 1998). Here, the well known clustering property of the Dirichlet process provides a nonparametric prior for the number of mixture components within each group, with the following generative model:

$$G \mid \alpha_0, G_0 \sim \text{DP}(\alpha_0, G_0) \qquad (1)$$
$$\theta_i \mid G \sim G \qquad \text{for each } i,$$
$$x_i \mid \theta_i \sim F(\theta_i) \qquad \text{for each } i,$$

where $G_0$ is a base measure of the DP, $\alpha_0 \geq 0$ is a concentration parameter, $\text{DP}(\cdot, \cdot)$ is a Dirichlet process, and $\theta_i$ is a parameter drawn from $G$. Each datum $x_i$ is then drawn from a distribution $F(\cdot)$ parameterized by $\theta_i$. In our models and experiments hereafter, $F(\cdot)$ is a Gaussian density. Ferguson (1973) showed that draws from a DP are discrete with probability one, and therefore, in a sufficiently large data set, there will be several $i$ for which the $\theta_i$ are identical; this gives rise to the natural clustering phenomenon giving DP mixtures their name.

Now, consider several groups, $J$, of data, denoted $((x_{ji})_{i=1}^{n_j})_{j=1}^{J}$; in such a setting it is natural to consider sets of Dirichlet processes, one for each group, and we still desire to tie the mixture models in the various groups. We therefore consider a hierarchical model, specifically one in which each of the child Dirichlet processes in the set has a base measure that is itself distributed according to a Dirichlet process. Such a base measure being again discrete with probability one, the child Dirichlet processes necessarily share atoms, thus ensuring that the the mixture models in the different groups necessarily share mixture components. The generative model is given by:

$$G_0 \mid \gamma, H \sim \text{DP}(\gamma, H) \qquad (2)$$
$$G_j \mid \alpha_0, G_0 \sim \text{DP}(\alpha_0, G_0) \qquad \text{for each } j,$$
$$\theta_{ji} \mid G_j \sim G_j \qquad \text{for each } j \text{ and } i,$$
$$x_{ji} \mid \theta_{ji} \sim F(\theta_{ji}) \qquad \text{for each } j \text{ and } i.$$

Figure 1(b) shows the graphical model, where there is a plate not only over data, $n$, but also over the $J$ (non-overlapping) groups of data. In this model the number of mixture components is unknown a priori for each group, and also for the data as a whole.

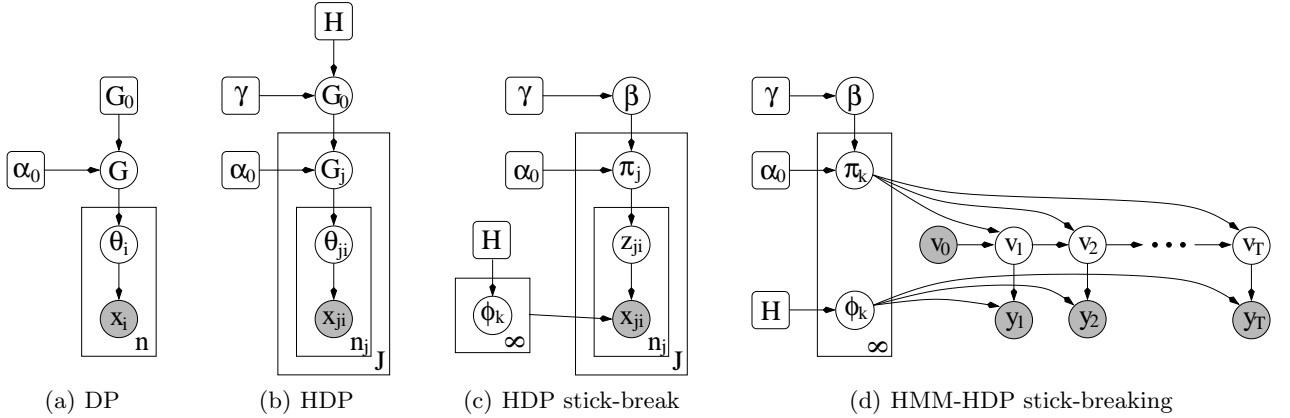

(a) DP     (b) HDP     (c) HDP stick-break     (d) HMM-HDP stick-breaking

Figure 1: Graphical model descriptions of the HDP-HMM as it compares to the HDP: **(a)** a DP mixture model; **(b)** the original HDP model; **(c)** the stick-breaking interpretation of the original HDP: mixing proportions for the $j$th group are drawn from common weights $\boldsymbol{\beta}$, themselves drawn from a stick-breaking process, and then each item $\boldsymbol{x}_{ji}$ is drawn from a mixture model with mixing proportions $\boldsymbol{\pi}_j$. **(d)** An unraveled HDP graphical model, wherein the next-state distribution is a current-state-dependent DP mixture model; each item $\boldsymbol{y}_t$ is drawn from a mixture model (having mixture indicator $v_t$) with mixing proportions $\boldsymbol{\pi}_{v_{t-1}}$ determined by the previous hidden state.

## 2.2 Recasting the Hidden Markov Model

The most straightforward route to understanding the connection between a hidden Markov model and the HDP described above is to first realize the stick-breaking characterization (Sethuraman 1994) of the HDP, given below and depicted in Figure 1(c),

$$\boldsymbol{\beta} \mid \gamma \sim \text{Stick}(\gamma) \qquad (3)$$
$$\boldsymbol{\pi}_j \mid \alpha_0, \boldsymbol{\beta} \sim \text{DP}(\alpha_0, \boldsymbol{\beta}) \qquad z_{ji} \mid \boldsymbol{\pi}_j \sim \boldsymbol{\pi}_j$$
$$\phi_k \mid H \sim H \qquad x_{ji} \mid z_{ji}, (\phi_k)_{k=1}^\infty \sim F(\phi_{z_{ji}}) ,$$

where the stick-breaking construction (Stick) gives rise to weights $\boldsymbol{\beta} = (\beta_k)_{k=1}^\infty$, with

$$\beta'_k \sim \text{Beta}(1, \gamma) \qquad \beta_k = \beta'_k \prod_{l=1}^{k-1}(1 - \beta'_l) . \qquad (4)$$

The advantage of this representation is that it makes explicit the generation of one (countably infinite) set of parameters $(\phi_k)_{k=1}^\infty$; the $j$th group has access to various of these parameters should it need them to model its data $(x_{ji})_{i=1}^{n_j}$, depending on the sampled mixing proportion $\boldsymbol{\pi}_j$.

Recall that a hidden Markov model (HMM) is a doubly stochastic Markov chain in which a sequence of multinomial "state" variables $(v_1, v_2, \ldots, v_T)$ are linked via a state transition matrix, and each element $y_t$ in a sequence of "observations" $(y_1, y_2, \ldots, y_T)$ is drawn independently of the other observations conditional on $v_t$ (Rabiner 1989). This is essentially a dynamic variant of a finite mixture model, in which there is one mixture component corresponding to each value of the multinomial state. Note that the HMM involves not a single mixture model, but rather a set of mixture models: one for each value of the current state. That is, the "current state" $v_t$ indexes a specific row of the transition matrix, with the probabilities in this row serving as the mixing proportions for the choice of the "next state" $v_{t+1}$. Given the next state $v_{t+1}$, the observation $y_{t+1}$ is drawn from the mixture component indexed by $v_{t+1}$.

Thus, to consider a nonparametric variant of the HMM which allows an unbounded set of states, we must consider a set of DPs, one for each value of the current state. Moreover, these DPs must be linked, because we want the same set of "next states" to be reachable from each of the "current states." This amounts to the requirement that the atoms associated with the state-conditional DPs should be shared—exactly the framework of the hierarchical DP. Thus, we simply replace the set of conditional finite mixture models underlying the classical HMM with an HDP, and the resulting model, an HDP-HMM, provides an alternative to methods that place an explicit parametric prior on the number of states or make use of model selection methods to select a fixed number of states (e.g. Stolcke and Omohundro 1993).

While there exist two Gibbs sampling methods for the HDP—one based on an extension of the Chinese Restaurant Process (CRP Aldous 1985) called the Chinese Restaurant Franchise (CRF), and the other based on an auxiliary variable method (both described in Teh et al. 2006)—only the auxiliary variable method is straightforward to implement for our HDP-HMM, and is used in our experiments shown below. In fact, in work that served as inspiration for the HDP-HMM (Beal et al. 2002), a sampler was presented that resembles the CRF scheme, but was necessarily ap-

proximate to reduce the time complexity of inference.[1]

The two-level urn model presented in that earlier work can be related to the HDP framework, by describing the latter using the stick-breaking formalism. In particular, consider the *unraveled* hierarchical Dirichlet process representation shown in Figure 1(d). The parameters in this representation have the following distributions, c.f. (3):

$$\begin{aligned} \boldsymbol{\beta} \mid \gamma &\sim \text{Stick}(\gamma) \\ \boldsymbol{\pi}_k \mid \alpha_0, \boldsymbol{\beta} &\sim \text{DP}(\alpha_0, \boldsymbol{\beta}) \quad v_t \mid v_{t-1}, (\boldsymbol{\pi}_k)_{k=1}^\infty \sim \boldsymbol{\pi}_{v_{t-1}} \\ \phi_k \mid H &\sim H \quad y_t \mid v_t, (\phi_k)_{k=1}^\infty \sim F(\phi_{v_t}), \end{aligned} \quad (5)$$

where we assume for simplicity that there is a distinguished initial state $v_0$. If we now consider the CRF representation of this model, it turns out that the result is equivalent to the coupled urn model of Beal et al. (2002). The advantage of this representation is that we can use an auxiliary variable sampling scheme that was designed for the HDP. As described above, the current instantiation, and most importantly the order, of the state variables $(v_1, \ldots, v_T)$, defines a grouping of the data $(y_1, \ldots, y_T)$ into groups indexed by the previous-state. Given this grouping, the settings of $\boldsymbol{\beta}$ and $\alpha_0$, the group-specific DP mixtures can be sampled independently. On sampling, the indicators $(v_1, \ldots, v_T)$ change, and hence the grouping of the data changes. Thus, given that we have access to a countably infinite set of hidden states, the HDP-HMM can be thought of as an HDP with an ever-shifting, countably infinite number of groups (metaphor: countably infinite tables in each of countably infinite restaurants, all sharing choices of dishes).

Last, as in the HDP, the HDP-HMM has hyperpriors on the hyperparameters $\alpha_0$ and $\gamma$, both gamma distributed with shape $a$. and inverse scale $b$. like so: $\alpha_0 \sim \text{Gamma}(a_{\alpha_0}, b_{\alpha_0}))$, $\gamma \sim \text{Gamma}(a_\gamma, b_\gamma)$; we sample over $\alpha_0$ and $\gamma$ during the auxiliary variable sampling scheme, so these are integrated out of the model.

## 3 Analysis of Experiments

### 3.1 Data & Sequence Similarity measure

We used two publicly available sets for the analysis: **i) (Iyer)** the gene expression time course data of Iyer et al. (1999), consisting of 517 genes' expressions across 12 time points; the expressions are log-normalized and standardized to have log expression 1 at time $t=1$ for all genes. Each gene is labeled belonging to a cluster 1–10 and a further outlier cluster is denoted as cluster '-1'. We assume that these labels are the result of biological expert modification following a preliminary Eisen (simple correlation in $T$ dimensions) analysis (Eisen et al. 1998). **ii) (Cho)** the second data set is the expression data described in Cho et al. (1998) consisting of 386 genes' expression across 17 time points, similarly normalized, with hand-labeled genes into one of 5 clusters.

We compare our HDP-HMM model to the standard HMM, referred to as finite HMM here after. For the finite HMM, we ran experiments with 7 different seed values and averaged over the various scores (explained below), in order to minimize the effects of initialization in EM. We define the (probabilistic measure of) dissimilarity between two genes' time courses for finite HMM as the $(c, d)$th element of a matrix $P$ (e.g. of size 517×517 for Iyer), which is the probability that the two time courses of each gene have identical hidden state trajectories. This can be computed straightforwardly after an E step in the Baum-Welch algorithm. Denoting the posterior over the hidden state at time $t$ of the $c$th gene sequence by $p(v_t^{(c)}|y_{1:T}^{(c)}, \Theta)$, where $\Theta$ are the current parameters of the HMM, then $\log P_{cd}$ is straightforwardly given by

$$\sum_{t=1}^T \log \sum_{r=1}^k p(v_t^{(c)} = r|y_{1:T}^{(c)}, \Theta) p(v_t^{(d)} = r|y_{1:T}^{(d)}, \Theta), \quad (6)$$

and therefore $P_{cd}$ ($=P_{dc}$) measures the probability of two genes, $c$ and $d$, having traversed similar entire hidden trajectories. We use $-\log P_{cd}$ as a measure of divergence (to be thought of as clustering distance) between genes $c$ and $d$.[2]

An analogous measure of divergence, or dissimilarity, can be computed for the HDP-HMM. In an infinite model, the posterior distribution over hidden state trajectories is represented as a set of samples, and the above quantity can be calculate simply from an empirical computation over the samples of trajectories taken over very long MCMC runs. Since the posterior samples always consist of represented hidden states, we do not suffer from the countably infinite state space. A similar method is used in the (i.i.d.) clustering using infinite mixture of Gaussians work of Rasmussen (2000) and Wild et al. (2002), but here we have extended to be a measure of similarity over *sequences*.

---

[1]There, bookkeeping for the CRF representation is very difficult: Adopting the well used CRP metaphor of customers entering a restaurant, in the CRF representation we have multiple restaurants, and sampling the table that a customer $\boldsymbol{y}_t$ sits at will influence the *restaurant* in which the following customer $\boldsymbol{y}_{t+1}$ must dine, and so on, resulting in a highly coupled system.

[2]More thoroughly one should compute a similarity that involves pairwise marginals over time as well, which would require a dynamic programming computation of similarity, which for the HDP-HMM is current research. However, we found the above approximation sufficient in our experiments. Note that the marginals used in (6), $p(v_t^{(c)} = r|y_{1:T}^{(c)}, \Theta)$, are still obtained using forward-backward.

### 3.2 Agreement with provided labels

We have used common external and internal indices to assess the quality of clustering obtained by various methods. Refer to the Appendix for more details on the definition of these metrics (note that DB$^*$ is the only score for which smaller values are better). We also use a recently introduced index called *purity*, as defined in Wild et al. (2002) which is a tree-based metric. Given a measure of dissimilarity between genes—provided by simple Eisen correlation, finite HMMs or HDP-HMMs—we can construct a dendrogram (we use *average* linkage), and on this binary-tree based representation we can fix a number of clusters, $C$, by severing the tree at this point, and force a putative labeling of each of the genes into each of $C$ possible labels.

### 3.3 Results

We compare simple correlation (Eisen, no time depedence) analysis to both finite HMMs of sizes varying from $k=1,\ldots,40$ and with the HDP-HMM with several settings of its hyper-hyperparameters. For the HDP-HMM, the auxiliary variable Gibbs sampling consisted of 100,000 burnin samples, collecting 250 posterior samples thereafter having a spacing of 750 samples between each.

In Tables 1 and 2, we display a subset of the indices as we vary $C$ for the Iyer data set only. Here, $b_\gamma$ denotes the hyper-hyperparameter setting over $\gamma$ for the HDP-HMM, and $k$ denotes the number of hidden states used in the finite HMM run. Noting the trade-off between sensitivity and specificity, and also the *in*difference to $b_\gamma$, we decided that for comparative purposes we would fix $C=11$ for Iyer. For the Cho results (not shown here for space), we used $C=5$, for similar reasons.

Table 2 shows a comparison of Eisen, finite HMM and HDP-HMM for various results of external and internal indices. A better visualization is given in Figure 2 for some of the Iyer results (Cho results are more impressive but are omitted due to space) in Figure 2. For clarity, we highlight entries in each column that perform the best, and reiterate that for DB$^*$ index lower is better (we do not consider the result of the finite HMM with $k=1$ since this degenerates to the case where all genes are in the same class ($-\log P_{cd}=0 \, \forall c,d$).

We note several trends: First, the degree of variation of all indices for different settings of the hyper-hyperparameter $b_\gamma$ is very small, suggesting that at this level of the Bayesian hierarchy the setting of the priors does not influence the learning of the model. Moreover, for any index, the variation over $k$ of the finite HMM is much larger. Second, it is clear that considering time in the HMM—especially in infinite

Table 1: Choice of number of clusters, $C$, for Iyer data.

|                      | rand | crand | jacc | spec | sens |
|----------------------|------|-------|------|------|------|
| $b_\gamma=1$, $C=1$  | 0.16 | 0.00  | 0.16 | 0.16 | 1.00 |
| 5                    | 0.72 | 0.35  | 0.33 | 0.35 | 0.89 |
| 10                   | 0.73 | 0.36  | 0.34 | 0.36 | 0.88 |
| 11                   | 0.73 | 0.36  | 0.34 | 0.36 | 0.88 |
| 15                   | 0.80 | 0.41  | 0.36 | 0.42 | 0.70 |
| 20                   | 0.82 | 0.38  | 0.33 | 0.43 | 0.57 |
| $b_\gamma=3$, $C=1$  | 0.16 | 0.00  | 0.16 | 0.16 | 1.00 |
| 5                    | 0.66 | 0.28  | 0.29 | 0.30 | 0.90 |
| 10                   | 0.73 | 0.35  | 0.33 | 0.35 | 0.83 |
| 11                   | 0.75 | 0.36  | 0.33 | 0.36 | 0.82 |
| 15                   | 0.80 | 0.39  | 0.35 | 0.41 | 0.69 |
| 20                   | 0.80 | 0.39  | 0.35 | 0.42 | 0.67 |

models—is advantageous compared to the simple correlation analysis of Eisen (already established Schliep et al. 2005). Third, there is evidence that the finite HMM is overfitting in the Iyer data set according to the Sil and DB indices, and for Cho according to several of the external indices; the HDP-HMM does not fit any one particular model, but integrates over a countably infinite set of models. Fourth, it is clear from the Table 2 that the vast majority of highlighted "winners" are for the HDP-HMM (last 7 rows), which shows a dramatic success over the finite HMM and Eisen (time independent) analyses.

*Inferred architecture:* Generally speaking the finite models show no improvement in performance beyond around $k=10$ hidden states. It is interesting, therefore, that in Figure 3(a) we find that for a wide range of $b_\gamma$ settings the HDP-HMM uses in excess of $k=20$ represented classes; the reason for this is that the architectures of the countably infinite and finite models are quite different, as shown in the transition matrices of Figures 3(b) and 3(c). The HDP-HMM has almost three times as sparse connectivity as its finite counterpart, with many states having only one or two possible destination states.

## 4 Conclusion and further directions

We have described the infinite HMM in the framework of the HDP, in which an auxiliary variable Gibbs sampling scheme is feasible. The HDP-HMM is in fact a CRF with a countably infinite number of tables in a countably infinite number of restaurants, all potentially sharing common dishes. We have shown for two time course gene expression data sets that the HDP-HMM performs similarly and, for most scenarios, better than the best finite HMMs found by model selection. We find that HMMs outperform the standard Eisen analysis based on simple correlation of the $T$-dimensional sequence vector, which treats the time points as independent. We used common measures of external and internal indices, including a tree-based in-dex, purity. We also find that the HDP-HMM models are learning quite different architectures for the hidden state dynamics. In current work we are examining more closely the prevalent paths through the hidden states, which may elucidate interesting regulatory networks at play; we will report the biological significance of these in a forthcoming article.

Table 2: Effect of varying the complexity $k$ of the finite models, and varying the hyper-hyperparameter $b_\gamma$.

| dataset | Iyer et al. (1999) ($C=11$) | | | | | | | | | Cho et al. (1998) ($C=5$) | | | | | | | | |
|---|---|---|---|---|---|---|---|---|---|---|---|---|---|---|---|---|---|---|
| index | rand | crand | jacc | sens | spec | sil | dunn | DB* | puri | rand | crand | jacc | sens | spec | sil | dunn | DB* | puri |
| Eisen | 0.80 | 0.38 | 0.33 | 0.63 | 0.41 | 0.55 | 1.542 | 0.70 | 0.58 | 0.77 | 0.43 | 0.41 | 0.68 | 0.51 | 0.37 | 1.357 | 0.78 | 0.58 |
| $k=1$ | 0.20 | 0.01 | 0.17 | 0.99 | 0.17 | $\infty$ | $\infty$ | 0.00 | 0.48 | 0.25 | 0.00 | 0.23 | 0.99 | 0.23 | $\infty$ | $\infty$ | 0.00 | 0.47 |
| 2 | 0.58 | 0.26 | 0.29 | 0.86 | 0.32 | 0.27 | 1.358 | 0.76 | 0.42 | 0.73 | 0.33 | 0.35 | 0.63 | 0.44 | 0.22 | 1.201 | 0.92 | 0.54 |
| 3 | 0.75 | 0.34 | 0.32 | 0.72 | 0.37 | 0.49 | 1.445 | 0.56 | 0.47 | 0.62 | 0.22 | 0.31 | 0.74 | 0.34 | 0.44 | 1.606 | 0.70 | 0.49 |
| 4 | 0.77 | 0.37 | 0.34 | 0.73 | 0.40 | 0.53 | 1.685 | 0.53 | 0.52 | 0.63 | 0.21 | 0.29 | 0.68 | 0.34 | 0.40 | 1.610 | 0.70 | 0.48 |
| 5 | 0.77 | 0.38 | 0.35 | 0.76 | 0.41 | 0.50 | 1.329 | 0.62 | 0.52 | 0.65 | 0.27 | 0.33 | 0.73 | 0.38 | 0.40 | 1.540 | 0.72 | 0.55 |
| 6 | 0.75 | 0.36 | 0.34 | 0.78 | 0.38 | 0.51 | 1.427 | 0.63 | 0.53 | 0.70 | 0.30 | 0.33 | 0.65 | 0.41 | 0.41 | 1.548 | 0.75 | 0.54 |
| 7 | 0.76 | 0.39 | 0.36 | 0.81 | 0.40 | 0.54 | 1.389 | 0.63 | 0.54 | 0.72 | 0.35 | 0.37 | 0.69 | 0.44 | 0.44 | 1.381 | 0.72 | 0.56 |
| 8 | 0.75 | 0.36 | 0.34 | 0.78 | 0.38 | 0.53 | 1.518 | 0.62 | 0.53 | 0.72 | 0.33 | 0.35 | 0.64 | 0.44 | 0.50 | 1.767 | 0.65 | 0.54 |
| 14 | 0.71 | 0.34 | 0.33 | 0.84 | 0.36 | 0.59 | 1.625 | 0.55 | 0.56 | 0.67 | 0.30 | 0.35 | 0.74 | 0.40 | 0.56 | 1.538 | 0.57 | 0.58 |
| 20 | 0.73 | 0.36 | 0.34 | 0.82 | 0.38 | 0.57 | 1.583 | 0.58 | 0.57 | 0.65 | 0.30 | 0.35 | 0.82 | 0.38 | 0.61 | 1.343 | 0.55 | 0.60 |
| 30 | 0.74 | 0.36 | 0.34 | 0.84 | 0.38 | 0.54 | 1.449 | 0.60 | 0.55 | 0.65 | 0.30 | 0.35 | 0.82 | 0.38 | 0.60 | 1.445 | 0.56 | 0.59 |
| 40 | 0.75 | 0.38 | 0.35 | 0.82 | 0.39 | 0.55 | 1.509 | 0.62 | 0.55 | 0.66 | 0.31 | 0.36 | 0.80 | 0.39 | 0.63 | 1.421 | 0.54 | 0.59 |
| $b_\gamma=0.25$ | 0.77 | 0.37 | 0.34 | 0.74 | 0.38 | 0.53 | 1.511 | 0.57 | 0.54 | 0.74 | 0.40 | 0.41 | 0.77 | 0.46 | 0.61 | 1.885 | 0.59 | 0.60 |
| 0.5 | 0.78 | 0.40 | 0.36 | 0.80 | 0.39 | 0.54 | 1.510 | 0.58 | 0.60 | 0.80 | 0.46 | 0.42 | 0.63 | 0.56 | 0.44 | 1.578 | 0.68 | 0.60 |
| 1 | 0.73 | 0.36 | 0.34 | 0.88 | 0.36 | 0.62 | 1.436 | 0.47 | 0.52 | 0.73 | 0.39 | 0.40 | 0.77 | 0.45 | 0.62 | 1.948 | 0.59 | 0.61 |
| 2 | 0.77 | 0.38 | 0.35 | 0.80 | 0.38 | 0.55 | 1.525 | 0.55 | 0.58 | 0.79 | 0.46 | 0.42 | 0.66 | 0.54 | 0.51 | 1.512 | 0.60 | 0.62 |
| 3 | 0.75 | 0.36 | 0.33 | 0.82 | 0.36 | 0.59 | 1.788 | 0.54 | 0.54 | 0.73 | 0.40 | 0.40 | 0.78 | 0.46 | 0.63 | 1.853 | 0.50 | 0.60 |
| 4 | 0.80 | 0.42 | 0.37 | 0.76 | 0.42 | 0.54 | 1.878 | 0.60 | 0.56 | 0.71 | 0.36 | 0.38 | 0.76 | 0.43 | 0.46 | 1.574 | 0.71 | 0.59 |
| 5 | 0.78 | 0.40 | 0.36 | 0.80 | 0.39 | 0.51 | 1.374 | 0.63 | 0.60 | 0.72 | 0.38 | 0.39 | 0.79 | 0.44 | 0.63 | 2.189 | 0.51 | 0.61 |

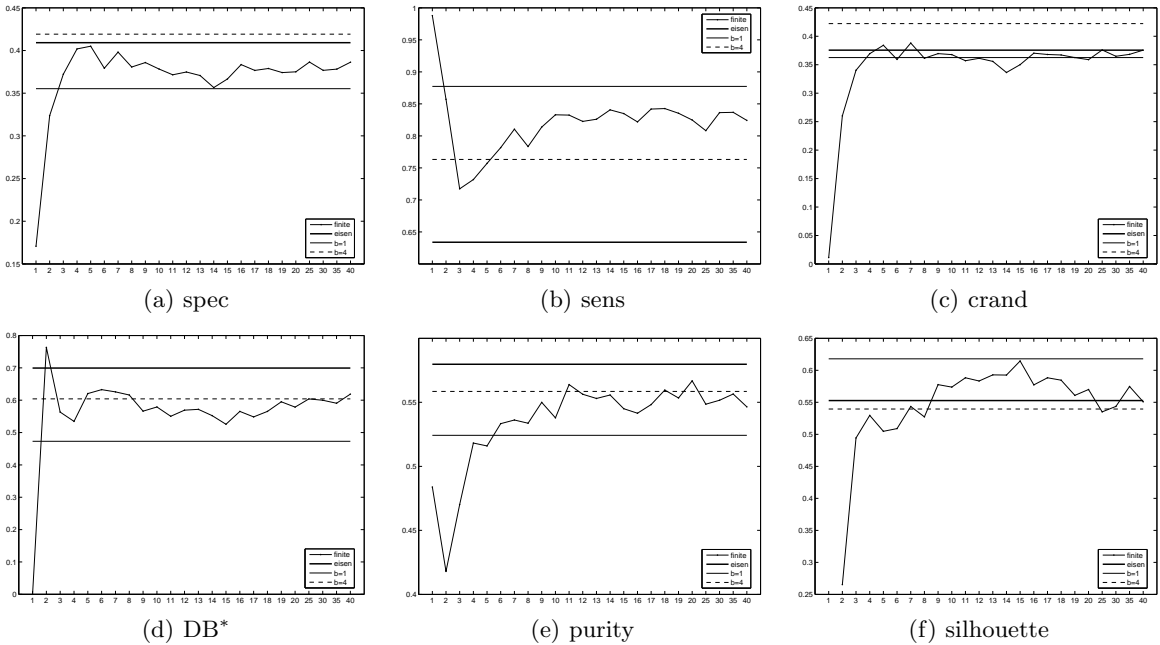

Figure 2: Relative performance of finite ($k=1,\ldots,40$, horizontal axis), infinite ($b_\gamma=1,4$, horizontal solid and dashed lines), and standard Eisen (correlation only, horizontal thick line) algorithms on a subset of the indices given in Table 2, for a cluster number $C=11$, for the Iyer data set.

In this paper we have not considered a leave-one-out analysis—either in terms of classification or in terms of density estimation—of a single gene using a model trained on the remainder. This is partly because sampling to calculate test likelihoods in the HDP-HMM is very time-consuming. However, previous preliminary work (Teh et al. 2005) on a simpler case of learning sequences of letters forming sentences on the *Alice's Adventures in Wonderland* showed that perplexity of test sentences was minimized using the HPD-HMM, as compared to Maximum Likelihood trained HMMs, Maximum A Posteriori trained HMMs, and variational Bayesian HMMs (Beal 2003). Also, Medvedovic and Sivaganesan (2002) show robustness of their infinite i.i.d. mixtures of Gaussians model as compared to finite mixtures, which is another reason to expect our time series analysis to perform well on such analyses.

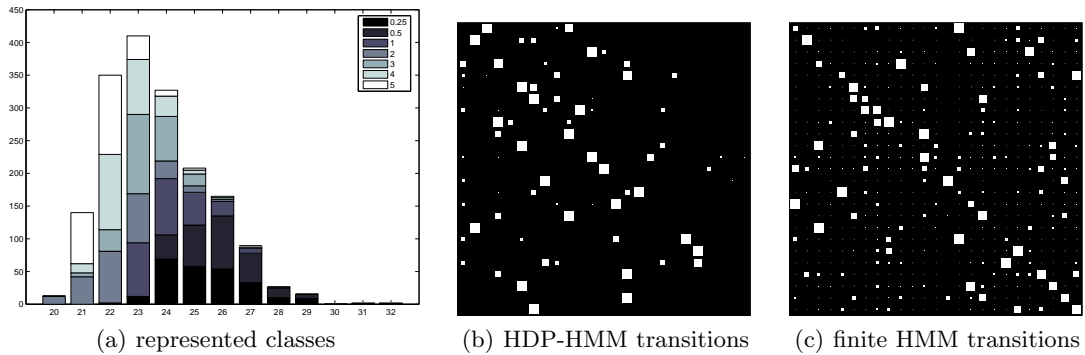

Figure 3: Analysis of hidden states in HDP-HMM (for Iyer data). **(a)** Distribution of number of represented classes in the HDP-HMM models. Shown are stacked values for various values of the (hyper-)hyperparameter $b_\gamma = (0.25, 0.5, 1, 2, 3, 4, 5)$, demonstrating that the mass over number of represented classes does not shift dramatically even over orders of magnitude of this high level hyperparameter; **(b)** Equivalent transition matrix for HDP-HMM ($b_\gamma = 0.25$, 11.7% non-zero entries): each *row* is a source state and its entries are transition probabilities that sum to 1, and brighter squares denote higher probability; **(c)** Transition matrix of the same size for the finite model—evidently less sparse (29.4% non-zero entries).

Finally, there are a host of exciting variants to the HDP-HMM. There are nested group models, which may be useful for capturing ontological information, and we are working on a countably infinite switching state-space model, as well as variants on the DP mixture model formalism in terms of Pitman-Yor processes that may have attractive properties in the domain of time-series modeling.

## Acknowledgements

We acknowledge support from NSF award 0524331, Yee Whye Teh for HDP code and discussion, and the helpful comments of anonymous reviewers.

# Appendices

Cluster validation is usually done by computation of indices, which signiify the quality of clustering, based either on a comparison with "ground-truth" labels (external) or without such comparison and relying on inherent qualities of the dendrogram or putative labels (internal).

## A.1 External indices

Let $n$ be the number of data points, $F' = \{F_1, ..., F_m\}$ be the possible labels, let $C' = \{C_1, ..., C_n\}$ be the clustering obtained by a clustering algorithm. Define two $n \times n$ incidence matrices, $F$ and $C$, s.t. $F_{ij} = 1$ if both the $i^{th}$ point and the $j^{th}$ point belong to same cluster in $F'$, 0 otherwise. And, $C_{ij} = 1$, if both the $i^{th}$ point and the $j^{th}$ point belong to same cluster in $C'$, 0 otherwise. Defining the following categories: $SS = \delta(C_{ij}, 1)\delta(F_{ij}, 1)$, $DD = \delta(C_{ij}, 0)\delta(F_{ij}, 0)$, $SD = \delta(C_{ij}, 1)\delta(F_{ij}, 0)$, and $DS = \delta(C_{ij} = 0)\delta(F_{ij}, 1)$, we use the following indices:

1. Rand index: $Rand = \frac{|SS|+|DD|}{|SS|+|SD|+|DS|+|DD|}$ .

2. Jaccard coefficient: $Jaccard = \frac{|SS|}{|SS|+|SD|+|DS|}$ .

3. CRand index; (C)orrected for chance assignments:
$CRand =$

$$\frac{\sum_{i=1}^{m}\sum_{j=1}^{n}\binom{n_{ij}}{2} - \binom{N}{2}^{-1}\sum_{i=1}^{m}\binom{n_i}{2}\sum_{j=1}^{n}\binom{n_j}{2}}{\frac{1}{2}\left[\sum_{i=1}^{m}\binom{n_i}{2}\sum_{j=1}^{n}\binom{n_j}{2}\right] - \sum_{i=1}^{m}\binom{n_i}{2}\sum_{j=1}^{n}\binom{n_j}{2}}$$

where $n_{ij}$ is the number of points in $P_i$ and $C_j$, $n_i$ is the number of points in the $P_i$, and $n_j$ the number of points in $C_j$.

4. The usual definitions for Sensitivity and Specificity:
$$Sens. = \frac{|SS|}{|SS|+|SD|} , \qquad Spec. = \frac{|SS|}{|SS|+|DS|} .$$

## A.2 Internal indices

Internal indices are computed to quantitatively assess clustering in the absence of provided labels, and attempt to evaluate *cohesion* (how similar are points in same clusters), and *separation* (how dissimilar are points in different clusters). Usually these indices are computed with a Euclidean distance metric, but to preserve the integrity of our analysis we used the $-\log P_{cd}$ dissimilarity given by (6).

**Silhouette:** For a given cluster, $C_j$ $(j = 1, ..., m)$, this method assigns to each sample of $C_j$ a quality measure, $s(i)$ $(i = 1, ..., n)$, known as the Silhouette width. The Silhouette width is a confidence indicator on the membership of the $i^{th}$ sample in cluster $C_j$, defined as

$$s(i) = (b(i) - a(i)) \ / \ \max\{a(i), b(i)\}$$

where $a(i)$ is the average distance between the $i^{th}$ sample and all of the samples in $C_j$, and $b(i)$ is the minimum average distance between the $i^{th}$ sample and all samples in $C_k$ $(k = 1, ..., c; k \neq j)$. For a given cluster, $C_j$ $(j = 1, ..., m)$, it is possible to calculate a cluster Silhouette $S_j = n^{-1}\sum_{j=1}^{n} s(i)$, which characterizes the heterogeneity and isolation properties of such a cluster, and a Global Silhouette value, $GS_u = m^{-1}\sum_{j=1}^{m} S_j$.

**Dunn's:** This index identifies sets of clusters that are compact and well separated. For any partition, U, produced by a clustering algorithm, let $C_i$ represent the $i^{th}$ cluster, the Dunn's validation index, D, is defined as:

$$D(U) = \min_{1 \leq i \leq m}\left\{\min_{\substack{1 \leq i \leq m \\ j \neq i}}\left\{\frac{\Delta'(C_i, C_j)}{\max_{1 \leq k \leq m}\{\Delta(C_k)\}}\right\}\right\},$$

where $\Delta'(C_i, C_j)$ defines the distance between clusters $C_i$ and $C_j$ (intercluster distance); $\Delta(C_k)$ represents the intracluster distance of cluster $C_k$, and $m$ is the number of clusters of partition. The main goal of this measure is to maximize intercluster distances whilst minimizing intra-cluster distances; large values of D correspond to good clusters.

**Davies-Bouldin:** This index is defined as:

$$DB(U) = \frac{1}{m}\sum_{i=1}^{c}\max_{j \neq i}\left\{\frac{\Delta(C_i) + \Delta(C_j)}{\delta(C_i, C_j)}\right\} ,$$

where $U$, $\delta(C_i, C_j)$, $\Delta(C_i)$, $\Delta(C_j)$ and $m$ are defined as in equation (7). Small values of DB correspond to clusters that are compact, and whose centers are far away from each other. Therefore, smaller DB is preferred.